\documentclass[runningheads]{llncs}
\usepackage{graphicx}
\usepackage{graphicx}
\usepackage{amsmath}
\usepackage{amssymb}
\usepackage{booktabs}
\usepackage[table,xcdraw]{xcolor}
\usepackage{longtable}
\usepackage{footmisc}
\usepackage{comment}
\usepackage{caption}
\usepackage{soul}
\usepackage{subcaption}
\usepackage{tikz}
\usepackage{comment}
\usepackage{color}
\usepackage{xspace}
\usepackage{wrapfig}
\usepackage{floatrow}
\usepackage{hyperref}
\usepackage[ruled]{algorithm2e}

\definecolor{lavenderblue}{rgb}{0.9, 0.9, 1.0}
\definecolor{whitee}{rgb}{1.0, 1.0, 1.0}

\usepackage{pifont}
\newcommand{\xmark}{\ding{55}}%

\newcommand{\eg}{e.g.\@\xspace} 

\newcommand{\ie}{i.e.\@\xspace}

\renewcommand{\appendixname}{
\begin{center}
\textbf{\Large Supplementary Materials: SCAM! Transferring humans between images with Semantic Cross Attention Modulation}
\end{center}
}

\newcommand{\mname}{SCAM\@\xspace}
\newcommand{\todo}[1]{\textcolor{red}{[todo: #1]}}

\newcommand{\nicolas}[1]{\textcolor{blue}{[#1]}}
\usepackage[accsupp]{axessibility}  

\begin{document}
\pagestyle{headings}
\mainmatter
\def\ECCVSubNumber{575}  

\title{SCAM! Transferring humans between images with Semantic Cross Attention Modulation} 

\titlerunning{SCAM}
%
\author{Nicolas Dufour\inst{1, 2}\orcidID{0000-0002-1903-5110} \and
David Picard\inst{1}\orcidID{ 0000-0002-6296-4222 } \and
Vicky Kalogeiton\inst{2}\orcidID{0000-0002-7368-6993}}
\authorrunning{Dufour et al.}
%
\institute{LIGM, Ecole des Ponts, Univ Gustave Eiffel, CNRS, Marne-la-Vallée, France 
\email{\{nicolas.dufour, david.picard\}@enpc.fr}
\and
LIX, CNRS, Ecole Polytechnique, IP Paris
\email{vicky.kalogeiton@lix.polytechnique.fr}
\\
Project page: \url{https://imagine.enpc.fr/~dufourn/scam}
}
\maketitle

\begin{abstract}
A large body of recent work targets semantically conditioned image generation.  
Most such methods focus on the narrower task of pose transfer and ignore the more challenging task of subject transfer that consists in not only transferring the pose but also the appearance and background. 
In this work, we introduce \mname (Semantic Cross Attention Modulation), a system that encodes rich and diverse information in each semantic region of the image (including foreground and background), thus achieving precise generation with emphasis on fine details. 
This is enabled by the Semantic Attention Transformer Encoder that extracts multiple latent vectors for each semantic region, and the corresponding generator that exploits these multiple latents by using semantic cross attention modulation. 
It is trained only using a reconstruction setup, while subject transfer is performed at test time.
Our analysis shows that our proposed architecture is successful at encoding the diversity of appearance in each semantic region. 
Extensive experiments on the iDesigner, CelebAMask-HD and ADE20K datasets show that \mname outperforms competing approaches; moreover, it sets the new state of the art on subject transfer.
\end{abstract}

\section{Introduction}
\label{sec:intro}

Being able to perform subject transfer between two images is a key challenge for many applications, from post-processing in game or art industries to software addressing the needs of the public. For instance, in film industries, one could replace a stunt performer by the main actors, thus alleviating the need of finding a look-alike performer; hence, increasing the filmmakers freedom. Similarly, it could enable finishing a film when an actor is indisposed.

%

\begin{figure}[t]
    \centering
    \includegraphics[width=\columnwidth]{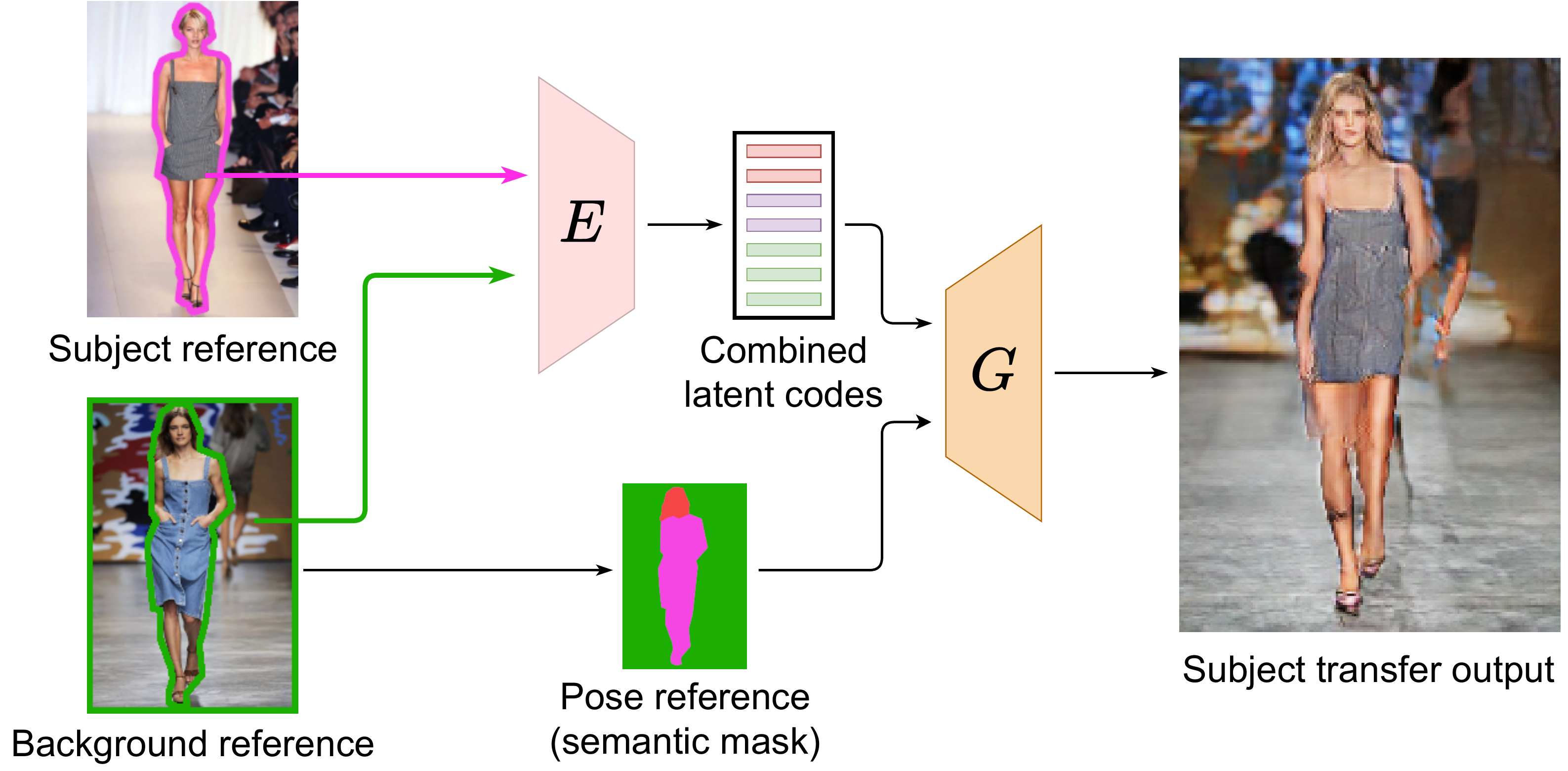}
    \caption[Overview of subject transfer by the proposed \mname.]{ 
    \small{
    \textbf{Subject transfer with the proposed \mname.} 
    We first encode the desired subject with the encoder E and get the subject latent codes. Then, we encode the background and the semantic mask for the pose and background reference. Finally, the generator G  synthesizes an image, where the subject is transferred with the desired background and pose. 
    %
    Pictures
    taken from the Internet\footnotemark}.
    }
    \label{fig:teaser}
\end{figure}

Given a source and a target subject, the idea of subject transfer is for the source subject to replace the target subject in the target image \emph{seamlessly}. The target image should keep the same background, the same interactions between subject and objects, and the same spatial configuration, to account for possible occlusions. Figure~\ref{fig:teaser} illustrates this. Note, in contrast to faces, buildings, or landscapes, human bodies are malleable with high morphological diversity, thus casting the task hard to model.


\footnotetext{Kate Moss picture by JB Villareal/Shoot Digital, Natalia Vodianova picture by Karl Prouse/Catwalking.}


Most methods focus either on pose transfer \cite{zhang2021pise,chan2019everybody,wang2018fewshotvid2vid}, where the pose changes, or on style transfer \cite{Zhu_2020,park2019semantic}, where the pose remains fixed but the subject's styling changes. 
These are limited and cannot be used out of the box for our task, as they are: 
(1) restrictive; they only work on uniform backgrounds, failing in complex ones (PISE~\cite{zhang2021pise}, SEAN~\cite{Zhu_2020},~\cite{wang2018fewshotvid2vid}), and 
(2) expensive; they require hard training~\cite{wang2018fewshotvid2vid} or training one model per human (Everybody Dance Now~\cite{chan2019everybody}). Instead, subject transfer changes both the pose and the style/identity of the subject. Thus, a successful system is decoupled in both pose and style transfer and performs both tasks simultaneously.

Semantic editing is a related task, consisting in controlling the output of a generative network by a segmentation mask. Indeed, subject transfer can be performed with semantic editing by using the mask of the target subject with the style of the source. 
However, modern methods cannot handle complex layout and rich structure (like full-bodies) with in the wild backgrounds. 
For instance, SPADE~\cite{park2019semantic} fails to control each region style independently, while SEAN~\cite{Zhu_2020} fails to handle complex, detailed scenes, such as multiple background objects.

To this end, we propose the \textbf{S}emantic \textbf{C}ross \textbf{A}ttention \textbf{M}odulation system (\textbf{\mname}), a semantic editing model that accounts for all aforementioned challenges relevant to subject transfer. 
\mname captures fine image details inside the semantic region by having multiple latents per semantic regions. This enables capturing unsupervised semantic information inside the semantic labels, which allows for better handling coarse semantic labels, such as  background. Our model can generate more complex backgrounds than previous methods and outperforms SEAN~\cite{Zhu_2020} both on the subject transfer and semantic reconstruction tasks.
 


We propose three architectural contributions: 
First, we propose the Semantic Cross Attention (\textbf{SCA}) that performs attention between a set of latents (each linked to a semantic region) and an image feature map. SCA constrains the attention, such as the latents only attend the regions on the image feature map that correspond to the relevant semantic label. 
Secondly, we introduce the \textbf{SAT} operation and encoder (\textbf{S}emantic \textbf{A}ttention \textbf{T}ransformer) that relies on cross attention to decide which information to gather in the image and for which latent, thus allowing for richer information to be encoded. 
Third, we propose the \textbf{SCAM}-Generator (after which \mname is named) that modulates the feature maps using the SCAM-Operation, which allows every pixel to attend to the semantically meaningful latents. Note, the whole architecture is trained using a reconstruction setup only, and subject transfer is performed at test time.


\section{Related Work}
\label{sec:relatedwork}

\noindent \textbf{Image to Image synthesis with GANs.} 
GANs~\cite{goodfellow2014generative} generate images by processing a random vector sampled from a predefined distribution with a dedicated network~\cite{lapgan,dcgan,karras2018progressive}. A major improvement is StyleGAN~\cite{karras2019stylebased,karras2020analyzing} that allows to modulate the feature map at each resolution according to a given style vector. 
Typically, unconditional GANs allow for minimal control over the generator's output. For more flexibility, Pix2Pix~\cite{isola2018imagetoimage,wang2018pix2pixHD} trains a generator coupled to an encoder, allowing for output control with multiple modalities (sketches, keypoints). One of its drawbacks is the need for data pairs, which can be hard to collect (drawings).  
To tackle this, CycleGAN~\cite{zhu2020unpaired} uses unpaired data by enforcing cycle consistency across domains. 
However, acquiring paired data is feasible when leveraging external models, such as semantic masks from segmentation models. In our case, we do not have access to ground-truth images where the subject has been transferred as it would require both the subject and the reference to have exactly the same pose and occlusion. 
We circumvent this by training on a reconstruction proxy task and performing subject transfer at test time.

\noindent \textbf{Semantic Image Generation.} Even if Pix2Pix~\cite{isola2018imagetoimage,wang2018pix2pixHD} manage to control the output image with a segmentation mask, it suffers from a semantic information washing-up. SPADE~\cite{park2019semantic} propose to fix this problem by introducing layer-wise semantic conditioning. CLADE\cite{tan2021efficient} propose a more efficient version of SPADE to reduce runtime complexity. Other approaches such as \cite{wang2021image,liu2019learning,li2021collaging,sushko2021need,tang2020local,tan2021efficient,endo2020diversifying,gu2019mask} propose improvements over spade. However, these approaches work well generating images from semantic information, but they do not focus on the case where we want to re-generate a given image to then be able to edit it. To do so, the style of the image must be carefully encoded and move from the single style vector used in SPADE and its variants. SEAN\cite{Zhu_2020} propose to introduce a single style vector per semantic label by performing average pooling on the encoder CNN features. GroupDNet~\cite{zhu2020semantically} propose to solve this by encoding each semantic region separately using grouped convolution. INADE~\cite{tan2021diverse} propose to use instance conditioned convolution to extract a single style code per instance. Although this allows better control of the output and richer representation, it still has two problems: 
(1) it is limited when handling coarse labels with diverse objects, (2) it creates a single vector per semantic region. 
In our approach, \mname, we solve this by introducing the SAT-Encoder which can extract rich representation from images and is able to output multiple and diverse latents per semantic region.
In turn, the SEAN-Generator modulates the output by both the semantic mask and the extracted style. This, however, modulates each pixel of a semantic region with the same style vector. 
Instead, our SCAM-Generator, uses attention to leverage different tokens to interact with, leading to different modulations per pixel, and hence enabling the emergence of unsupervised semantic structure in the semantic regions.
Other approaches propose to use diffusion process approaches \cite{meng2022sdedit} to perform this editing process. However, these diffusion approaches are very expensive to sample from.

\noindent \textbf{Attention in Computer Vision.} Despite their remarkable success~\cite{devlin2019bert,radford2018improving,radford2019language,brown2020language}, transformers suffer from a quadratic complexity problem, which makes it hard to use. 
To tackle this, most vision methods~\cite{dosovitskiy2021image,touvron2021training,liu2021swin} subdivide  images into patches, resulting in losing information. 
Instead, the recent Perceiver~\cite{jaegle2021perceiver,jaegle2021perceiverio} tackles this complexity issue by replacing self attention by cross attention. 
The image pixels are attended by a significatively smaller set of learned tokens.

\begin{figure*}[t!]
    \centering
    \includegraphics[width=\textwidth]{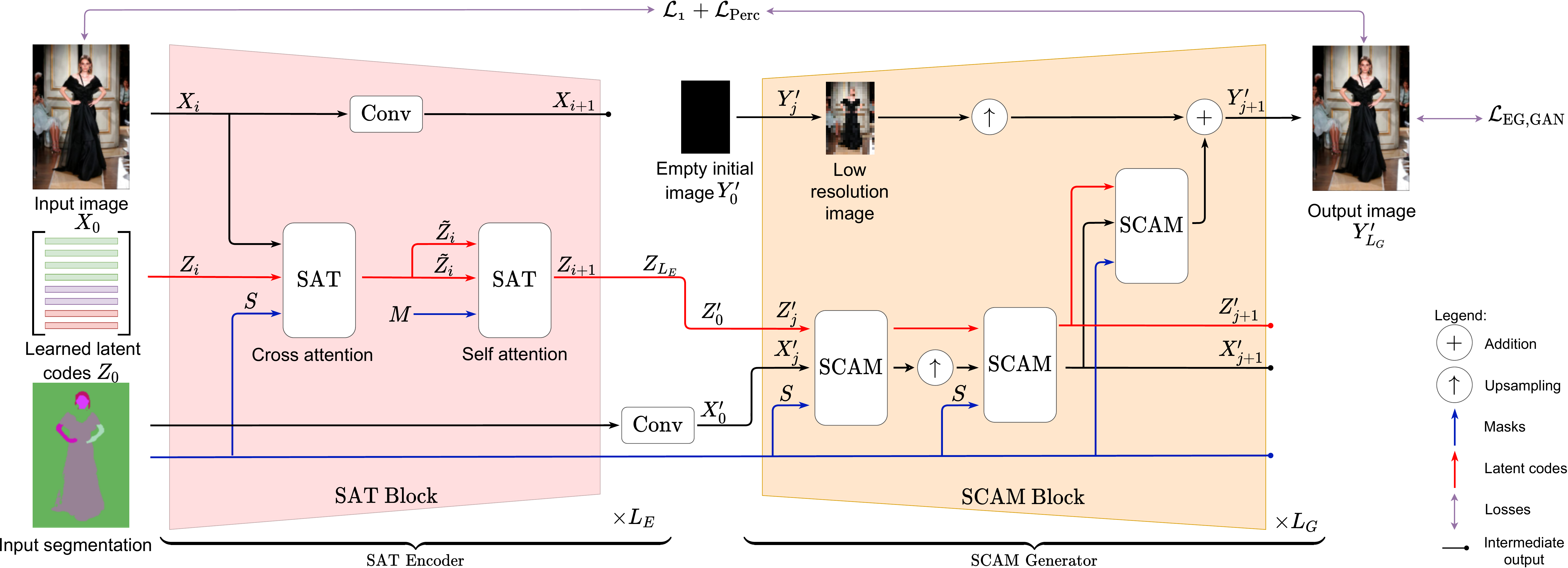}
     \caption{
     \small{\textbf{Training setup of the proposed \mname architecture}. 
     It consists of the SAT-Encoder (pink) and the SCAM-Generator (yellow). 
     The SAT-Encoder allows the latents to retrieve information from an image, exploiting both the raw image and the convolution feature maps. Once the image is encoded, the latents are fed to the SCAM-Generator, which captures top-down and bottom-up interactions with a semantic constraint, allowing to easily alter the desired regions thanks to the latents that are dedicated to a given region.
     }}
    \label{fig:arch}
\end{figure*}

\noindent \textbf{Attention in GANS.} It~\cite{zhang2019selfattention,esser2021taming,jiang2021transgan,zhao2021improved,lee2021vitgan,zhang2021styleswin} has shown great progress over the past years.
GANsformer~\cite{hudson2021generative} leverages cross attention to exploit multiple style codes between style vectors and feature maps. 
These  architectures use attention for unconditional generation; however, they do not focus on  subject transfer. 
Instead, our proposed SCA improves upon GANsformer's duplex attention for semantically constrained generation by assigning latents to semantic regions.

\noindent \textbf{Pose Transfer.} 
Using keypoints for pose transfer~\cite{ma2017pose,zhu2019progressive,tang2020xinggan,li2020pona} typically results in coarse representation of bodies. 
To tackle this, some methods use semantic masks~\cite{han2019clothflow,dong2018soft,zhang2020human,zhang2021pise}. 
These, however, focus only on pose transfer, which does not alter the background. Instead, we aim at subject transfer, where preserving the background is crucial. 
Most methods are limited to simple backgrounds. \cite{chan2019everybody} overfit a GAN to a video and regenerate the background; hence, it cannot be used for dynamic scenes. \cite{wang2018fewshotvid2vid} address this by adapting the weights of the generator, but this ties the subject to the background, not allowing for subject transfer. Here, we focus on subject transfer that changes both pose and background.

\section{Method}
\label{sec:methods}


Our goal is to perform semantic editing with a focus on subject transfer. 
We propose the \textbf{\mname} method (Semantic Cross Attention Modulation, Figure~\ref{fig:arch}).   
It relies on \textbf{SCA} (Semantic Cross Attention), \ie a novel mechanism that masks the attention according to segmentation masks, thus encoding semantically meaningful latent variables (Section~\ref{sub:sca}). 
\mname consists of: 
(a)  \textbf{SAT-Encoder} (Semantic Attention Transformer) that relies on cross attention to decide which information to gather in the image and for which latent (Section~\ref{sub:sat}); and  
(b)  \textbf{SCAM-Generator} (Semantic Cross Attention Modulation) that captures rich semantic information in an unsupervised way (Section~\ref{sub:scam}).


\noindent \textbf{Notation.}
Let $X\in\mathbb{R}^{n\times C}$ be the feature map with n the number of pixels, and C the number of channels. Let $Z\in\mathbb{R}^{m\times d}$ be a set of $m$ latents of dimension $d$ and $s$ the number of semantic labels. Each semantic label is attributed $k$ latents, such that $m=k\times s$. Each semantic label mask is assigned $k$ copies in $S{\in}\{0;1\}^{n \times m}$. $\sigma(.)$ is the softmax operation.

\noindent \textbf{Motivation.} Since many regions of the image have visually diverse content (\eg, the background), we propose to encode this varied information in several complementary latents. Motivated by the findings of Gansformers\cite{hudson2021generative}, we use attention to introduce both a constraint on which part of the image a latent can get information from and a competing mechanism between latents attending the same region so as to specialize them. Reciprocally, using duplex attention, we introduce the same strategy by limiting the latents that the feature map can attend to and introduce a competition between parts of a semantic region that can attend the same latent.

%

\subsection{Semantic Cross Attention (SCA)}
\label{sub:sca}

\noindent \textbf{Definition.} The goal of SCA is two-fold depending on what is the query and what is the key. Either it allows to give the feature map information from a semantically restricted set of latents or, respectively, it allows a set of latents to retrieve information in a semantically restricted region of the feature map. 
%
It is defined as:  

\begin{footnotesize}
\begin{equation}
    \text{SCA}(I_{1}, I_{2}, I_{3}) = \sigma\left(\frac{QK^T\odot I_{3} +\tau \left(1-I_{3}\right)}{\sqrt{d_{in}}}\right)V \quad ,
\end{equation}
\end{footnotesize}
\noindent where $I_{1},I_{2},I_{3}$ the inputs, with $I_{1}$ attending $I_{2}$, and $I_{3}$ the mask that forces tokens from $I_1$ to attend only specific tokens from $I_2$\footnote{{\label{fnote1}}The attention values requiring masking are filled with $-\infty$ before the softmax. (In practice $\tau{=}-10^9$) }, $Q {=} W_QI_{1}$, $K {=} W_KI_{2}$ and $V {=} W_VI_{2}$ the queries, keys and values, and $d_{in}$ the internal attention dimension.

We use three types of SCA. 
\noindent \textit{(a) SCA with pixels $X$ attending latents $Z$}: $\text{SCA}(X, Z, S)$, where $W_{Q} {\in} \mathbb{R}^{n\times d_{in}}$ and $W_{K}, W_{V} {\in} \mathbb{R}^{m\times d_{in}}$.
The idea is to force the pixels from a semantic region to attend latents that are associated with the same label. 
\noindent \textit{(b) SCA with latents $Z$ attending pixels $X$}: $\text{SCA}(Z, X, S)$, where $W_{Q}{\in} \mathbb{R}^{m\times d_{in}}$, $W_{K}, W_{V} {\in} \mathbb{R}^{n\times d_{in}}$. 
The idea is to semantically mask attention values to enforce latents to attend semantically corresponding pixels.
\noindent \textit{(c) SCA with latents $Z$ attending themselves}: $\text{SCA}(Z, Z, M)$, where $W_{Q}, W_{K}, W_{V} {\in} \mathbb{R}^{n\times d_{in}}$. We denote $M\in\mathbb{N}^{m\times m}$ this mask, with $M_{\text{latents}}(i,j) {=} 1$ if the semantic label of latent $i$ is the same as the one of latent $j$; $0$ otherwise.
The idea is to let the latents only attend latents that share the same semantic label. 
\subsection{SAT-Encoder}
\label{sub:sat}

Following~\cite{jaegle2021perceiver}, our SAT-Encoder relies on cross attention. 
It consists of $L_{E}$ consecutive layers of \textbf{SAT-Blocks}, where the input of the $i{+}1$-th one is the output of the $i$-th one 
(Figure~\ref{fig:arch}~(left)). Given a set of learned queries $Z_0$ (\ie, parameters updated with gradient descent using back-propagation), it 
outputs latents $Z_{L_{E}}$ that have encoded the input image. 
This allows to create multiple latents per semantic regions resulting in specialized latents for different part of each semantic region of the image. The encoder is also flexible enough to easily assign a different number of latent for each semantic region, allowing to optimize the representation power given to each semantic region. At each layer, the latent code retrieves information from the image feature maps at different scales.


\noindent The \textbf{SAT-Block} is composed of three components: two \textbf{SAT-Operations} and a strided convolution. SAT-Operations are transformer-like \cite{vaswani2017attention} operations, replacing self-attention by our proposed SCA. They are defined as:
\begin{align}
\text{SAT}(I_1, I_2, S) = \text{LN}( f( \text{LN}(\text{SCA}(I_1, I_2, S) + I_1)) + I_1),
\end{align}
with $LN$ the layer norm and $f$ a simple 2-layer feed forward network.
The first SAT-Operation, $\text{SAT}(X, Z, S)$, let the latents retrieve information from the image feature map (case \emph{(b)} from \ref{sub:sca}). The second SAT-Operation, $\text{SAT}(Z, Z, M)$, is refining the latents using SCA in a self attention setup where the latents attend themselves, keeping the semantic restriction (case \emph{(c)} from \ref{sub:sca}). The strided convolution encode the previous layer image feature map, reducing its spatial dimension. Implementation details and reference code about SAT-Blocks and SAT-Operation are in the supplementary material

\subsection{SCAM-Generator}
\label{sub:scam}


\noindent \textbf{Definition and Architecture.} 
SCAM-Generator takes as an input the latent codes $Z'_0$ ($=Z_{L_E}$ encoder's output) and the segmentation mask $S$ and outputs the generated image $Y'_{L_{G}}$. 
It consists of $L_{G}$ \textbf{SCAM-Blocks} (Figure~\ref{fig:arch}~(right)). 
The input latent of the generator is given by the encoder's output, whereas the input latents of each block within the generator are the output latents of previous blocks. 
Similarly, the input feature maps of each block are the feature map outputs at each resolution, while for the first features, we encode the segmentation with a convolutional layer following~\cite{park2019semantic,Zhu_2020}.

\noindent The \textbf{SCAM-Block} has a progressive growing architecture similar to the one of StyleGAN2 \cite{karras2020analyzing} and consists of 3 SCAM-Operations (See Figure~\ref{fig:arch}~(right)). 2 SCAM-Operations process the generator feature-map, with an upscaling operation between the two, while a parallel SCAM-operation retrieves information from the feature maps and generate the image in the RGB space. Implementation details and reference code are in the supplementary.




\begin{figure}[b]
\floatbox[{\capbeside\thisfloatsetup{capbesideposition={right,center},capbesidewidth=4cm}}]{figure}[\FBwidth]
{\caption{\small{\textbf{SCAM-Operation}. 
    It modulates a feature map according to a segmentation map, allowing each pixel to retrieve information from a semantically restricted set of latents. It enables both top-bottom (latents retrieve information from the feature map) and bottom-top interactions (the map gets information from latents).
    }}}
{\includegraphics[width=6.5cm]{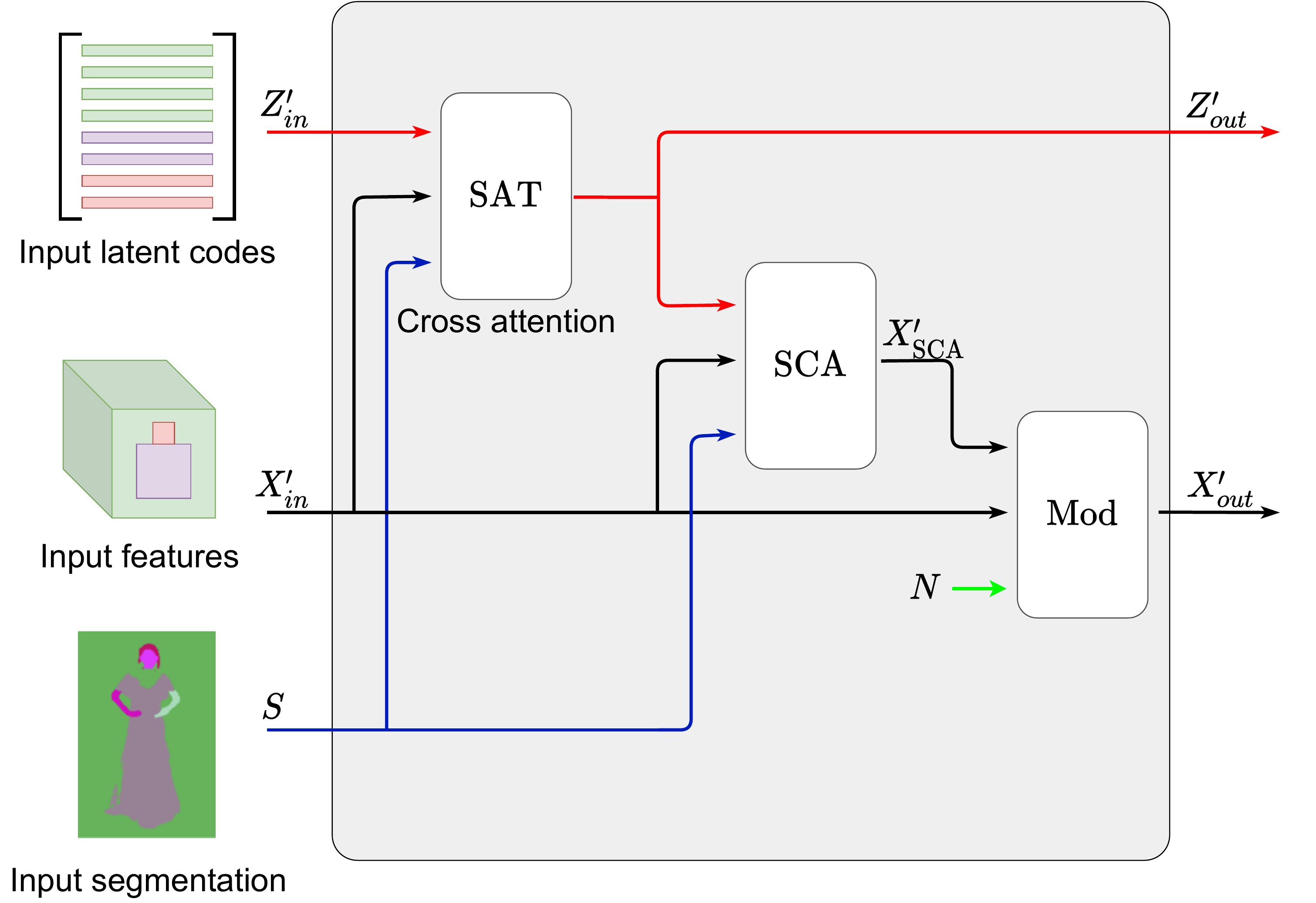}
\label{fig:scam}}
\end{figure}

\noindent \textbf{SCAM-Operation.} It aims at exchanging information between pixels/features and latents of the same semantic label and is depicted in Figure~\ref{fig:scam}. 
Each SCAM-Operation has inputs: (1) the set of input latents $Z'_{\text{in}}$,(2) the input feature map $X'_{\text{in}}$ , and (3) the segmentation mask $S$. 
Its outputs are: (1) the output latents $Z'_{\text{out}}$, and (2) the output feature map $X'_{\text{out}}$. 
SCAM-Operation consists of three parts: (a) the latent SAT, (b) the feature SCA, (c) the Modulation operation.

\noindent \textbf{(a) The latent SAT.} It uses a SAT operation to update the current latents   $Z'_{\text{in}}$ based on the current feature map: 
$Z'_{\text{out}}{:=} \text{SAT}(Z'_{\text{in}},X'_{\text{in}},S)$. 
This allows for latent refinement while enforcing the latents semantic constraint thanks to the SCA operation inside SAT.

\noindent \textbf{(b) The feature SCA} performs latent to image attention: it incorporates the latent information to the pixels/features  using SCA. Given $X'_{\text{in}}$, the output latents from the SAT-Operation $Z'_{\text{out}}$ and the mask $S$, it outputs  $X'_{\text{SCA}}$:
\begin{equation}
    X'_{\text{SCA}} = \text{SCA}(X'_{\text{in}},Z'_{\text{out}},S) \quad .
\end{equation}

\noindent \textbf{(c) The Modulation operation} takes as input the $X'_{\text{in}}$ maps and the  $X'_{\text{SCA}}$ from SCA, and outputs features that are passed through a convolution layer $g(.)$ to produce the final output image features $X'_{\text{out}}$.  It is defined as:
\begin{equation}
\begin{small}
    X'_{out} =g\left(\gamma(X'_{\text{SCA}})\odot \text{IN}(X'_{\text{in}}) + \mu(X'_{\text{SCA}}) + N\right) \quad .
    \end{small}
\end{equation}
It consists of the $\gamma(.)$ and $\mu(.)$ operators that determine the scale and bias factor for the modulation operation. 
We use Instance Normalization (IN)~\cite{ulyanov2017instance} to perform the feature map normalization. 
Following~\cite{karras2019stylebased,karras2020analyzing}, we also add noise $N {\in} \mathbb{R}^{n \times C}$, with $N {\sim}\mathcal{N}(0,\sigma^2 I_{n \times C})$, and $\sigma$ learned parameter. This encourages the modulation to account for stochastic variation.

\noindent \textbf{Discussion.} SCAM-Operation outputs both the modulated feature map and the updated latents. Thus, the information from the previous maps is propagated to the latents and to the feature maps. This brings several advantages: 
first, it preserves the semantic constraint; 
second, it provides finer refinement within the semantic mask by attending to multiple latents; and 
third it allows each pixel to choose which latent to use for modulation.

\subsection{Training losses}
We train \mname with GAN and reconstruction losses. We denote by D the discriminator, G the generator and E the encoder.
For the \textbf{ discriminator}, we follow~\cite{isola2018imagetoimage,wang2018pix2pixHD,wang2018pix2pixHD,park2019semantic} and use PatchGAN, as it discriminates for patches of the given image instead of the global image. 
For the \textbf{GAN loss}, we use Hinge GAN loss~\cite{lim2017geometric}: $\mathcal{L}_{\text{EG,GAN}}$, $\mathcal{L}_{\text{D,GAN}}$. 
For the \textbf{reconstruction loss}, we use the perceptual loss $\mathcal{L}_{\text{Perc}}$ as in~\cite{Zhu_2020} and the $\mathcal{L}_1$  between the input and the reconstructed input. 
The \textbf{final losses} are $\mathcal{L}_{\text{EG}} =  \mathcal{L}_{\text{EG,GAN}} + \lambda_{\text{perc}}\mathcal{L}_{\text{Perc}} + \lambda_{1}\mathcal{L}_{1}$ and $\mathcal{L}_{\text{D}} {=} \mathcal{L}_{\text{D,GAN}} $ with $\lambda_{\text{perc}}$ and $\lambda_{1}$ hyperparameters. 
We use $\lambda_{\text{perc}}{=}\lambda_{1}{=}10$ in our experiments. Training details are in the supplementary.

\subsection{Subject transfer}
Once \mname is trained for reconstruction, at test time we perform subject transfer. 
Given two images $X_{A},X_{B}$ with their respective segmentation masks $(S_A,S_B)$, we retrieve their latent codes as $Z_A$ and $Z_B$ using the SAT-Encoder. To transfer the subject from $X_B$ to the context of $X_A$, we create $Z_{\text{mix}}$, where the style codes related to the background come from $Z_A$ and the remainder codes come from $Z_B$. Then, we retrieve $Y'_\text{mix} {=} G(Z_{\text{mix}},S_B)$. See Figure \ref{fig:teaser}.


\section{Experiments}
\label{sec:experiments}

We now present experimental results for \mname . More results and ethical impacts are discussed in supplementary.

\noindent \textbf{Implementation details.}
We train all models for 50k steps, with batch size of 32 on 4 Nvidia V100 GPUs. We set $k{=}8$ latents per label of dim $d{=}256$. We generate images of resolution 256px.


\subsection{Datasets and metrics}


\noindent \textbf{iDesigner}~\cite{idesigner} contains images from designer fashion shows, including 50 different designers with 50k train and 10k test samples.
We segment human parts with~\cite{li2019selfcorrection} and then merge the labels to end up with: face, body and background labels.

\noindent \textbf{CelebAMask-HQ}~\cite{lee2020maskgan} contains celebrity faces from CelebA-HQ ~\cite{karras2018progressive} with 28k train and 2k test images labelled with 19 semantic labels of high quality. 

\noindent \textbf{ADE20K}~\cite{zhou2017scene} contains diverse scenes images with 20k train and 2k test images. The images are labelled with 150 semantic labels of high quality.

\noindent \textbf{Metrics.} We use PSNR, reconstruction FID (R-FID), and swap FID (S-FID). R-FID is computed as the FID between the train set and the reconstructed test set, while S-FID is between the test set and a set of subject transfer images computed on the test set. We introduce REIDAcc and REIDSim, computed in the latent space of a re-identification network~\cite{fu2021unsupervised}. REIDSim computes the average cosine similarity between the subject image and the subject transfer image. REIDAcc accounts for the proportion of images where the cosine similarity of the transferred subject is higher with the subject than with the background.

\subsection{Comparison to the state of the art}

\begin{table}[t]
\begin{center}
		\begin{small}
		\resizebox{\linewidth}{!}{
\begin{tabular}{l ccccc c ccc c cc}
\toprule
Method & \multicolumn{5}{c}{\cellcolor{lavenderblue}{\textbf{iDesigner}}} & & \multicolumn{3}{c}{\cellcolor{lavenderblue}{\textbf{CelebAMask-HQ}}}  & & \multicolumn{2}{c}{\cellcolor{lavenderblue}{\textbf{ADE20K}}} \\
& PSNR $\uparrow$ & R-FID $\downarrow$ & S-FID $\downarrow$ & REIDSim $\uparrow$  & REIDAcc $\uparrow$ & & PSNR $\uparrow$ & R-FID $\downarrow$ & S-FID $\downarrow$ & & PSNR $\uparrow$ & R-FID $\downarrow$  \\
\midrule
SPADE \cite{park2019semantic} [CVPR19] & 10.4   & 66.7  & 67.5 & 0.67 & 0.26 &  & 10.9 & 38.2 & 38.3 & & 10.7 & 59.7 \\ 
CLADE\cite{tan2021efficient}[TPAMI21] & 11.3 & 45.4  & 46.1 & 0.68 & 0.29 &  & 10.8 & 41.8 & 42.0 & & 10.4 & 53.7 \\ 
SEAN-CLADE \cite{tan2021efficient} [TPAMI21] & 15.3   & 48.4  & 56.1 & 0.75 & 0.31 &  & 16.2 & 19.8 & 24.3 & & 14.0 & 38.7 \\ 
INADE\cite{tan2021diverse} [CVPR21] & 12.0   & 33.0  & 33.9 & 0.72 & 0.34 &  & 12.24 & 22.7 & 23.4 & & 11.3 & 48.6 \\ 
SEAN \cite{Zhu_2020} [CVPR20] & 14.9  & 53.5 & 58.7 & 0.74 & 0.30 & & 16.2 & 18.9 & 22.8 & & 14.6 & 47.6 \\ 
\mname (Ours) & \textbf{21.4} & \textbf{13.2} & \textbf{26.9} & \textbf{0.81} & \textbf{0.56} & & \textbf{21.9} & \textbf{15.5} & \textbf{19.8} & & \textbf{20.0} & \textbf{27.5} \\

\bottomrule
\end{tabular}}
		\end{small}
	\end{center}
	\caption{\small{\textbf{Comparison} on iDesigner~\cite{idesigner} and CelebAMask-HQ~\cite{lee2020maskgan} and ADE20K\cite{zhou2017scene}.}}
\label{tab:quant-sota}
\end{table}

We first compare our proposed \mname to INADE~\cite{tan2021diverse}, SEAN~\cite{Zhu_2020}, CLADE~\cite{tan2021efficient}, SEAN-CLADE~\cite{tan2021efficient} and  SPADE~\cite{park2019semantic}. We reproduce all the methods and provide code for our implementations in the supplementary material.

\noindent \textbf{Results on iDesigner} are shown in 
Table~\ref{tab:quant-sota}~(left). 
Overall, \mname outperforms competing approaches for all metrics. 
Specifically, for PSNR it outperforms SEAN-CLADE by approximately +5.9dB, whereas it reaches 13.2 R-FID vs 33.0 for INADE.
These major boosts show that our reconstructed images  better preserve the details of the initial images, meaning the representation power of \mname is higher than that of other approaches. 
The difference is also notable for S-FID, with INADE reaching 33.9 vs 26.9 for \mname, showing that our method perform better subject transfer on datasets with coarse semantic labels than other approaches.
We also observe a superiority of SCAM on REIDSim (+0.06 compared to INADE) and REIDAcc (+0.22 compared to INADE).
Overall REIDAcc is a hard metric. This can be explained by the fact that the subject transfer image shares the same semantic information with the background image.

%

\noindent \textbf{Results on CelebAMask-HQ} are shown in Table~\ref{tab:quant-sota}~(center), where we observe that \mname outperforms all methods. 
For instance, for PSNR, \mname outperforms SEAN by +5.7dB. \mname also improves over SEAN by 3.4 R-FID points  (15.5 vs 19.8). 
For subject transfer, \mname outperforms SEAN by a S-FID decrease of almost 3 points (19.8 vs 22.8), clearly indicating that our method is also better at transferring. 
We observe that even for a dataset that has precise labelling, our approach still outperforms competing approaches.

\noindent \textbf{Results on ADE20K} are shown in Table~\ref{tab:quant-sota}~(right), where we observe that \mname outperforms all methods. 
\mname has the best PSNR of 20.0 whereas second to best SEAN has a PSNR of 14.6. \mname also beats SEAN-CLADE by 11.2 R-FID points  (27.5 vs 38.7). 
We cannot evaluate S-FID on this dataset since it is hard to select what is the main subject in the image, and not all images share the same semantic labels.

\begin{table*}[t]
\begin{minipage}[b]{\textwidth}

\centering
		\begin{small}
		\resizebox{\linewidth}{!}{
		\begin{tabular}{l cc c c c ccc c cc cc cc c cc cc cc}
\toprule
	
 & \multicolumn{2}{c}{\cellcolor{lavenderblue}{\textbf{Encoder}}} & & \multicolumn{1}{c}{\cellcolor{lavenderblue}{\textbf{Generator}}} & &
\multicolumn{3}{c}{\cellcolor{lavenderblue}{\textbf{Losses}}} & & 
\multicolumn{6}{c}{\cellcolor{lavenderblue}{\textbf{iDesigner}}} & & 
\multicolumn{6}{c}{\cellcolor{lavenderblue}{\textbf{CelebAMask-HQ}}} \\ 
& Conv & SA & & SAT& &$\mathcal{L}_1$ & $\mathcal{L}_{\text{Perc}}$ & $\mathcal{L}_{\text{GAN}}$ & & \multicolumn{1}{c}{PSNR} $\uparrow$ & $+\Delta$ & R-FID $\downarrow$ &$-\Delta$   &S-FID $\downarrow$ & $-\Delta$ & & \multicolumn{1}{c}{PSNR} $\uparrow$ & $+\Delta$ & R-FID $\downarrow$ &$-\Delta$   &S-FID $\downarrow$ & $-\Delta$ \\
\midrule

i	&	 \checkmark	& \checkmark & &	\checkmark &  	&	 \checkmark	&	\checkmark	&	\checkmark 	& &	21.0	&		&	13.2	&		&	27.1	& 	& &	22.0	&		&	15.7	&		&	20.7	& 	\\
ii 	&	\xmark & \checkmark	& &	\checkmark & 	&	\checkmark	&	\checkmark	&	\checkmark 	&	&	$19.2$	&	$-1.8$	&	$26.3$	&	$-13.1$   &	$34.1$	& 	$-7.0$	& &	$\mathbf{22.1}$ &	$+0.1$	&	$\mathbf{15.5}$	&	$+0.2$	&	$20.2$	&	$+0.5$ \\
iii	& \checkmark &	\xmark & & \checkmark & &	\checkmark	&	\checkmark	&	\checkmark 	&	&	$21.3$	&	$+0.3$	&	$\mathbf{12.7}$	&	$+0.5$	&	$\mathbf{26.1}$	& 	$+1.0$	& &	$22.0$	&	$0.0$ 	&	$16.5$	&	$-0.8$	&	$\mathbf{20.0}$	&	$+0.7$	\\
iv	& \xmark &	\xmark & & \checkmark & &	\checkmark &	\checkmark	&	\checkmark	& &	18.5	&	$-2.5$	&	$24.8$	&	$-11.6$	&	$29.8$	&	$-2.7$	&	& $19.8$	&	$-2.2$ 	&	$19.0$	&	$-3.3$	&	$21.6$	&	$-0.9$	\\
v	&	\checkmark &	\checkmark &	&	\xmark &	&	\checkmark	&	\checkmark	&	\checkmark 	& 	&	$21.1$	&	$+0.1$	&	$15.6$	&	$-2.4$	&	$27.7$	&	$-0.6$ 	&	& $21.8$	&	$-0.2$	&	$15.6$	&	$+0.1$	&	$21.7$	&	$-1.0$ 	\\
vi	&	\xmark &	\checkmark &	&	\xmark &	&	\checkmark	&	\checkmark	&	\checkmark 	& 	&	$17.7$	&	$-3.3$	&	$49.8$	&	$-36.6$	&	$55.9$	&	$-28.8$	& &	$20.2$	&	$-1.8$	&	$21.7$	&	$-6.0$	&	$25.3$	&	$-4.6$	\\
vii	&	\checkmark &	\xmark &	&	\xmark &	&	\checkmark	&	\checkmark	&	\checkmark 	& 	&	$21.0$	&	$0.0$	&	$16.7$	&	$-3.5$	&	$32.3$	&	$-5.2$	& &	$21.4$	&	$-0.6$	&	$16.7$	&	$-1.0$	&	$21.8$	&	$-1.1$ 	\\
viii	&	\checkmark  &	\checkmark & 	&	\checkmark & 	&	\xmark	&	\checkmark	&	\checkmark 	& 	&	$19.5$	&	$-1.5$	&	$16.1$	&	$-2.9$	&	$30.7$	&	$-3.6$ 	& &	-	&	-	&	-	&	-	&	-	&	- 		\\
ix	&	\checkmark  &	 \checkmark &	&	\checkmark 	& &	\checkmark 	&	\xmark	&	\checkmark 	& 	&	$\mathbf{22.9}$	&	$+1.9$	&	$43.2$	&	$-30.0$	&	$91.3$	&	$-64.2$ 	&	& -	&	-	&	-	&	-	&	-	&	- 	\\
\midrule
x	&	 \multicolumn{2}{c}{SEAN} &	& SCAM 	& &	\checkmark 	&	\checkmark	&	\checkmark 	& 	&	$17.5$	&	$-3.5$	&	$27.6$	&	$-14.4$	&	$32.7$	&	$-5.6$ 	& &	$17.6$	&	$-2.4$	&	$20.4$	&	$-5.3$	&	$24.2$	&	$-3.5$ 	\\
\bottomrule
\end{tabular}}
		\end{small}
	\caption{\small{\textbf{Ablation} study on iDesigner~\cite{idesigner} and  CelebAMask-HQ\cite{lee2020maskgan}} }
\label{tab:ablation-idesinger}

\end{minipage}
\end{table*}

\subsection{Ablations}

Here, we perform several ablations to validate the effectiveness of all components of \mname. We denote by $-$ experiments that do not converge.

\noindent \textbf{Ablations of SAT-Encoder and SCAM-Generator.}
We examine the effectiveness of our encoder and generator by modifying either the SAT-Encoder or the SCAM-Generator with baseline versions and report the results on iDesigner in Table~\ref{tab:ablation-idesinger}. The first row (i) corresponds to our \mname. 
We benchmark some variants of our SAT-Encoder: Conv denotes whether we use convolutions in SAT or not. SA denotes whether we use self-attention SAT block or not.
The results show that overall convolutions in the SAT-Encoder provide a big encoding advantage. This is especially true for the complex iDesigner dataset. Indeed (i) outperforms (ii, iv, vi) by a high margin on iDesigner e.g., by -13.1 R-FID and -7.0 S-FID for (ii), which validates our use of multiple resolutions in the encoder.
We also examine a variation of the generator by removing the SAT block in the SCAM block.
Having SAT in SCAM leads to better results; such as -2.4 in R-FID, and -0.6 in S-FID for (v) on iDesigner. Similar results can be observed in (vi and vii).

In (x), we use the same encoder as in SEAN and the SCAM generator. To manage to have multiple latents per semantic latents, we split the SEAN encoding in 8 smaller latents for each semantic latent. We observe here that our SAT-Encoder is better than the SEAN encoder at extracting information from images. Indeed, we obtain better R-FID and S-FID (-14,4/-5.6) with our encoder than the SEAN encoder.

\noindent \textbf{Ablation of Losses.} 
Table~\ref{tab:ablation-idesinger}~(i,viii,ix) ablates the three $\mathcal{L}_1,\mathcal{L}_{\text{Perc}},\mathcal{L}_{\text{GAN}}$ losses used in \mname.  
The full combination (i) reaches the best results. 
Interestingly, removing $\mathcal{L}_{\text{Perc}}$ (ix) results in the best PSNR point (22.9db) while having among the worst R-FID and S-FID (43.2, 91.3, respectively). 
This is expected, as removing the perceptual loss makes the generator rely only on the $L_1$ loss, and may artificially increase PSNR at the cost of realism.

\noindent \textbf{Number $k$ of latents per semantic label.} 
We've studied the impact of the number of latents $k$ per semantic label on the performance of the model. We tested $k\in[4,8,16,32]$ on iDesigner~\cite{idesigner}. We find that the R-FID increase with the number of latents where the S-FID decreases. Indeed for $k=4$ we have R-FID of 15.9 and a S-FID of 27.3. For $k=32$, we have  R-FID of 9.7 and a S-FID of 30.3. We settle for $k=8$ in our experiments since it offers the best trade-off between S-FID and R-FID. We also privilege a smaller k since the time complexity of our model is $O(k)$.



\begin{figure}[b]
\caption{\small{(a) \mname specializes latents on complex patterns (shirt, eyes, shoes); (b) it learns semantic information \emph{on its own} inside the semantic labels (background people).}}
\includegraphics[width=\textwidth]{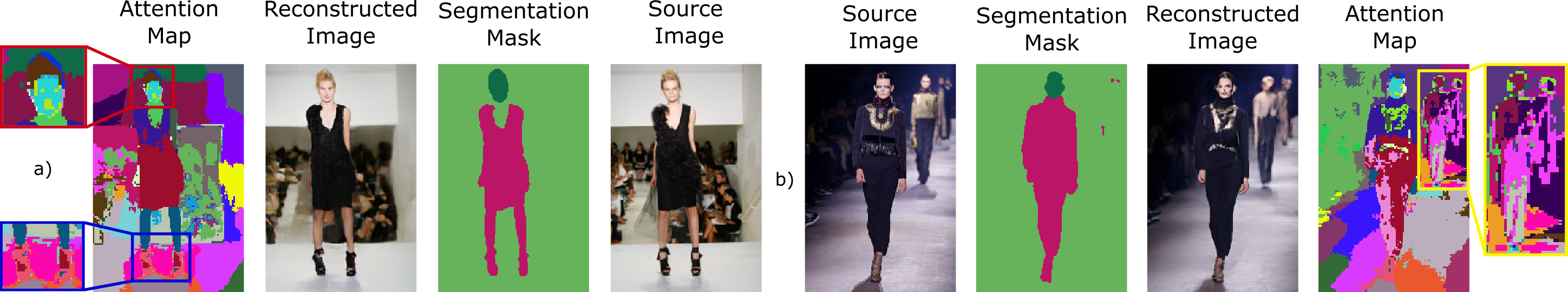}
\label{fig:attention_vis}
\end{figure}
\noindent \textbf{Visualization of the attention matrix.} 
To investigate how using multiple latents per region is handled by \mname, we visualize in Figure~\ref{fig:attention_vis} the last \mname layer attention matrix. We colour each pixel according to the corresponding latent with the highest attention value.
Overall, we observe that for each semantic region, the latents attend to different subregions, capturing semantic information without supervision. 
The first example (a) shows that even without specialized segmentation labels, \mname specializes some latents to reconstruct the complex face pattern (eyes, mouth, and hair) and others for the different body parts (dress and shoes). 
%
The second row displays an interesting case: 
\mname is capable of assigning different latents to the humans in the background even if they are not labelled as such.

\subsection{Qualitative results on reconstruction}

\begin{figure}[t] 
    \centering
    \includegraphics[width=\columnwidth]{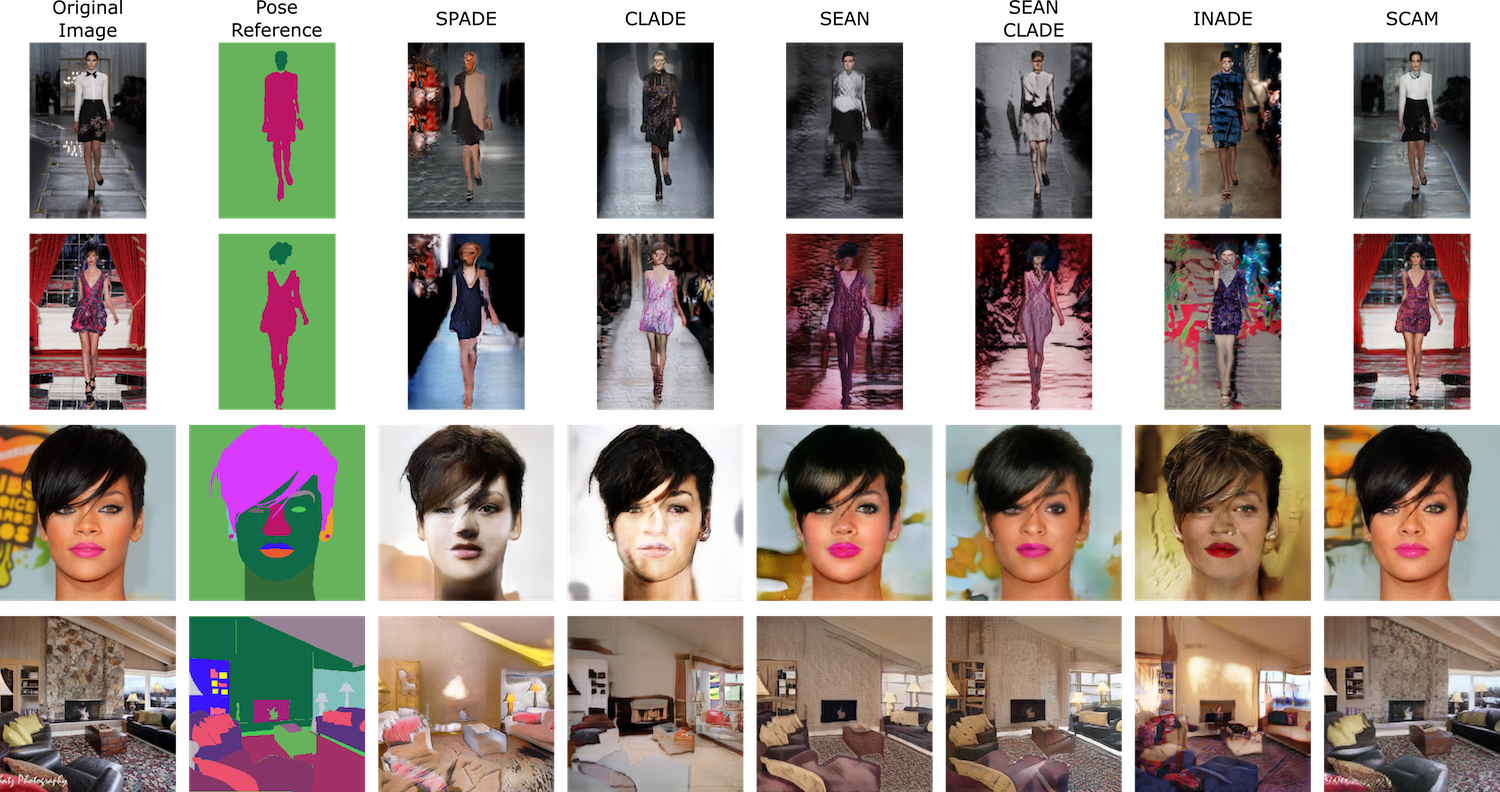}
    \caption{\small{
    \textbf{Reconstructions} on  iDesigner~\cite{idesigner}, CelebAMask-HQ\cite{lee2020maskgan} and ADE20K\cite{zhou2017scene}.}
    }
    \label{fig:reco}
\end{figure}

Overall, we observe that the reconstruction quality of \mname is superior to competing approaches. Figure~\ref{fig:reco} displays the input images, their masks (first two columns) and the reconstructions by SPADE~\cite{park2019semantic}, CLADE~\cite{tan2021efficient}, SEAN~\cite{Zhu_2020}, SEAN-CLADE~\cite{tan2021efficient}, INADE~\cite{tan2021diverse} and \mname (last six columns). The first two rows are samples from iDesigner, the third is from CelebAMask-HQ and the lastrow from ADE20K.

\noindent \textbf{Subject reconstruction on iDesigner.} \mname reconstructs more structure than SEAN, both in the background and the human. INADE, CLADE and SPADE tend to generate images that doesn't match the style of the original image. For the \textit{background}, we observe that the curtains and window frame of the second row are well-reconstructed by \mname, in contrast to SEAN that includes colors but no other frame-cues. This highlights the rich generation capabilities of SCAM-Generator, which manage to generate complex backgrounds where competing approaches fail. For the \textit{subject}, \mname results in finer reconstructions compared to other approaches. For instance, in the first row \mname reconstructs coherent clothes, while SEAN generate a blurred out version of the clothes.

\noindent \textbf{Reconstruction on CelebAMask-HQ.}  Overall, \mname generates crisper and more realistic results than competing approaches. For instance, in the third row, SEAN fails entirely to reconstruct the background by producing an averaged texture. Our method does a better job at this, figuring out a better positioning for the logo and capturing better colors and shapes. On the subject, we also observe that SEAN fails to capture small details such as eye colors.

\noindent \textbf{Reconstruction on ADE20K.}  \mname has a more reliable reconstruction than competing approaches. In the fourth row, we observe that \mname is the only approach that manage to reconstruct some texture on the chimney wall. We also observe that overall, \mname has more detail in the reconstructed object whereas approaches such as SEAN tend to have more averaged textures.

\subsection{Qualitative results on subject transfer}
\begin{figure*}[t] 
    \centering
    \includegraphics[width=\textwidth]{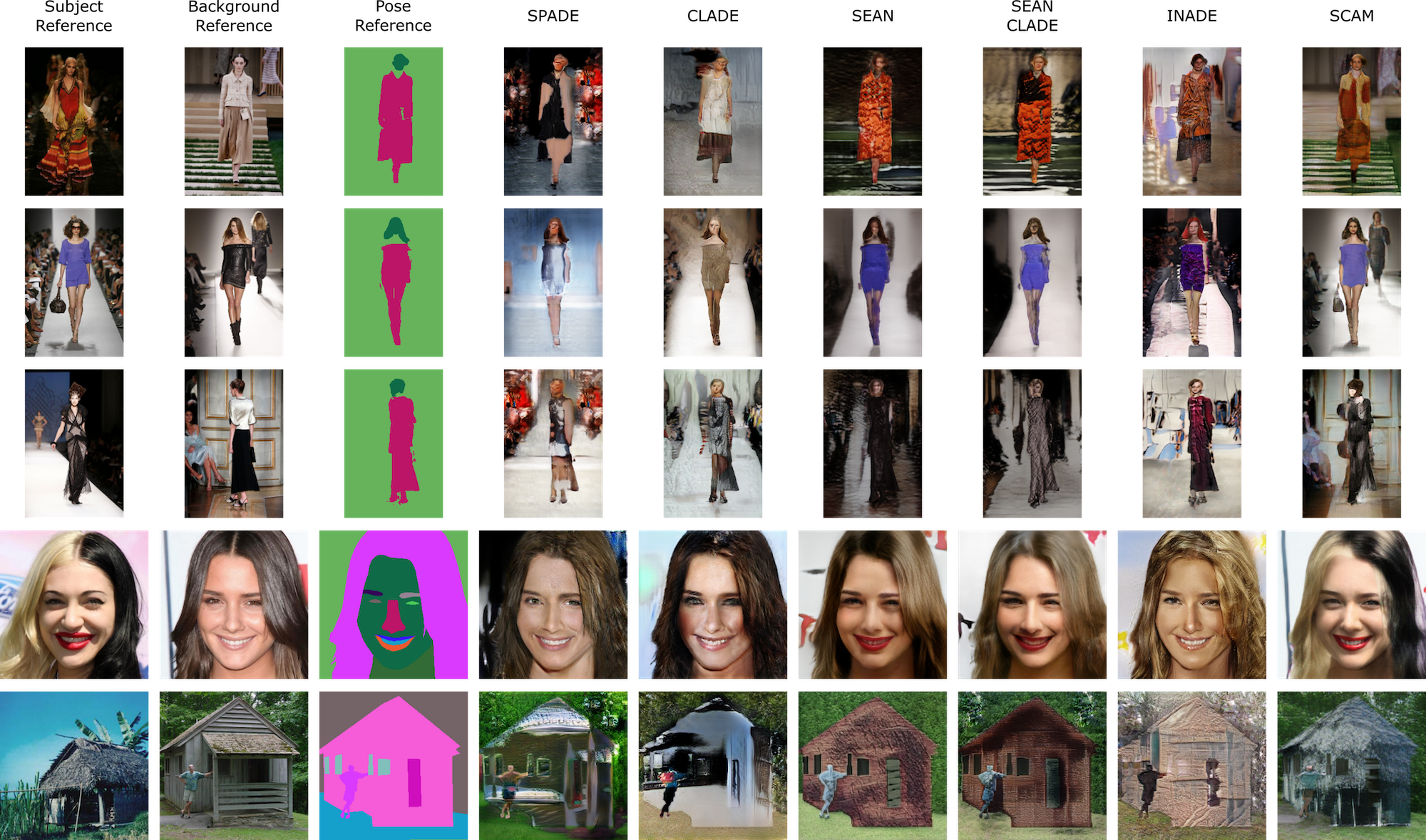}
    \caption{\small{\textbf{Subject Transfer} on the test set of iDesigner~\cite{idesigner}, CelebAMask-HQ\cite{lee2020maskgan} and ADE20K\cite{zhou2017scene}. Note the hard case in row 3, where only \mname rotates the subject. For ADE20K, we consider the house as the subject in the 5th row. 
    }}
    \label{fig:subject_transfer}
\end{figure*}
Figure~\ref{fig:reco} displays the subject and background images and the segmentation mask of the pose (first three columns). Then we show the subject transfer for SPADE~\cite{park2019semantic}, CLADE~\cite{tan2021efficient}, SEAN~\cite{Zhu_2020}, SEAN-CLADE~\cite{tan2021efficient}, INADE~\cite{tan2021diverse} and \mname (last six columns). The first three rows are samples from iDesigner, the fourth is from CelebAMask-HQ and the last one is on ADE20K.
Overall, amongst all methods, \mname is the one to successfully preserve all components of subject transfer: subject appearance, background appearance and pose.

\noindent \textbf{Subject transfer on iDesigner.} \mname leads to superior subjects (e.g., washed out colors in SEAN vs coherent structured clothing in \mname in the first row) and background reconstruction (i.e., global background structure, positioning of people on the sides and colours) than SEAN. For instance, in the first row, SEAN completely fails to reconstruct the background, while SEAN-CLADE reconstructs some texture but lacks detail. Similarly, in the second row, \mname captures the model in the background whereas other approaches miss it; note also the precise people reconstruction in the left part of the catwalk for \mname compared to competing approaches. 
In the third row, we have a very hard case, where the subject has a completely different pose than the pose reference. While other approaches fail to rotate the subject, \mname does succeed in doing, even if the quality of the generated image is low. Overall, subjects have better appearance in \mname than in other methods, like details in clothes, shoes, or faces.


\noindent \textbf{Subject transfer on CelebAMask-HQ.} The fourth row shows that \mname recovers more details in the transferred image, such as the colour of the skin or facial expression. Notably, \mname does capture the bicolor separation of the hair, while SEAN, SEAN-CLADE and INADE display an averaged hair color. 
\noindent \textbf{Subject transfer on ADE20K.} In the fifth row, we consider the house as the subject we want to transfer. We can see that SCAM does transfer the hut like appearance of the subject, whereas competing approaches fail to do so. We also observe that most approaches have difficulties with the person generation, only \mname generates a human that is coherent with the background reference.

\subsection{User Study on iDesigner}
We perform an user study on iDesigner. We compare 3 models one against each other: INADE, SEAN and SCAM. We have asked 38 different people to select among 20 reconstruction images which method in their opinion had the best reconstruction. We denote the percentage of best pick for each method as R-UP (Reconstruction User Preference). Similarly, we have the same people asked to select among 20 subject transfer images which method performed the best subject transfer in their opinion. Participant were asked to take into account image quality and quality of transfer. We denote the percentage of best pick for each method as ST-UP (Subject Transfer User Preference).

\noindent We observe that our method, SCAM, outperforms competing methods by a high margin. Indeed, 98.4\% of users chose SCAM as the best image reconstruction technique, 1.5\% picked SEAN and 0.1\% for INADE. As for subject transfer 92.8\% of users preferred SCAM, 5.0\% chose SCAM and 2.2\% INADE.

\section{Conclusion}
\label{sec:conclusion}
We introduced \mname that performs semantic editing and in particular subject transfer in images. The architecture contributions of \mname are: 
first, the semantic cross attention (\textbf{SCA}) mechanism  performing attention between features and a set of latents under the constraint that they only attend to semantically meaningful regions; second, the Semantic Attention Transformer Encoder (\textbf{SAT}) retrieving information based on a semantic attention mask; third, the Semantic Cross Attention Modulation Generator \textbf{(SCAM)} performing semantic-based generation. 
\mname sets the new state of the art by leveraging multiple latents per semantic region and by providing a finer encoding of the latent vectors both at encoding and decoding stages.
%

\noindent\textbf{Acknowledgments: }We would like to thank Dimitrios Papadopoulos, Monika Wysoczanska, Philippe Chiberre and Thibaut Issenhuth for proofreading and Simon Ebel for helping with the video. This work was granted access to the HPC resources of IDRIS under the allocation 2021-AD011012630 made by GENCI and 
was supported by a DIM RFSI grant and ANR project TOSAI
ANR-20-IADJ-0009.

\clearpage
%
%
\bibliographystyle{splncs04}
\bibliography{shortstrings,egbib}
\clearpage


\appendix
\appendixname


\paragraph{}
In this supplementary material, we first discuss the societal and environmental impact of \mname (Section~\ref{sup:sec:discussion}). 
Then, we give more details of \mname and additional experimental analysis (Sections~\ref{sup:sec:about}-\ref{sup:sub:exps}).

\section{Discussion}
\label{sup:sec:discussion}

\subsection{Societal impact}
This work could allow for multiple applications such as editing images, where we can easily exchange someone by someone else. This can have a negative impact if used with bad intentions. Indeed, one could use our method to create fake news. As of today, deep fakes can still be detected, by either humans or algorithms. However, as research progress in these fields, detection is going to get more complicated and regulation will have to control this.

Our method could also be very useful to create movies. It could allow changing an actor by another one. This could be very helpful for finishing a movie when an actor is impaired. It could also reduce the producing cost when using superstars, where an actor could accept for his image to be used, while not acting physically himself. This however could pose some legal problems since it is easy to create content without permission of the subject.

\subsection{Environmental impact}
For this project, we used 42.3 thousands GPUs hours. The experiments were done on a GPU cluster. The GPUs used are Nvidia V100-32g. Our experiments used 80\% of the GPUs maximum power of 250Wh, which amounts to 10.6 MWh of energy used for the whole project. The cluster we used is the Jean Zay cluster, situated in France. France heavily relies on nuclear energy, having a greener energy than average, with 50-80g CO2 for each kWh produced. Considering only the CO2 for the production of the electricity used, this results in 528-846kg of CO2 emitted for this project. Training a single SCAM model takes around 50 GPUs hours, which amounts to 10kWh and 500-800 g of CO2 emitted. As a comparison, the world average per capita CO2 emission is 4.7 ton/year.
\subsection{Future Work}
To improve upon this method, we see three main directions. First, this project is intended for images. For it to work well on videos, we would need to add some temporal information. Adding temporal consistency and smoothing should help have more visually pleasing outputs. Second, another improvement would be to directly train on transferred images. In this project, we train on the reconstructed images, and then we infer subject transfer. We conjecture that training on the transferred images would yield better results. Third, we observe that the subject transfer task is complicated by the quality of the segmentation masks and the disparity between the pose reference subject segmentation and the style reference subject. To prevent this, we could learn a network to map the pose reference subject segmentation to the wanted pose.

\section{About SCAM}
\label{sup:sec:about}

\subsection{Implementation details}
We train all models for 50k steps, with a batch size of 32 on 4 Nvidia V100 GPUs. We use AdamW~\cite{loshchilov2017decoupled} with a learning rate of 0.0001 for the generator and the encoder, and 0.0004 for the discriminator with $\beta {=} (0.9,0.999)$. 
We set $k{=}8$ latents per label of dimension $d{=}256$.
Each time we use attention on a feature map, we first encode the image using 2D sinusoidal positional encodings, as attention acts on sets that do not include positional information. Note that adding positional encoding to the latents is not useful because latents can be seen as a set and not a sequence. Moreover, since the latents are going to be initialized from a learned vector, this learned vector can incorporate the positional information, if needed, and this information will propagate throughout the architecture.
In the discriminator, we use GradNorm~\cite{wu2021gradient} to stabilize the training. For the SAT-Encoder, we have $L_E=6$ SAT-Blocks. In the SAT-Generator, we have $L_G=7$ SCAM-Blocks.

\subsection{Details on SAT}
Here are the detailed implementations for SAT-Operation (Cross (See Algorithm\ref{alg:sat_cross}) and Self (See Algorithm \ref{alg:sat_self})) and for the SAT Block (See Algorithm \ref{alg:sat_block}). For even finer details (Hyperparameters, practical details) see the provided Pytorch code.

\SetKwComment{Comment}{/* }{ */}

\begin{algorithm}
\caption{SAT-Cross at layer i}\label{alg:sat_cross}
\KwData{$Z_{i}\in\mathbb{R}^{m\times d}$;$X_{i}\in\mathbb{R}^{n\times C_i}$;$S\in\mathbb{1}^{ n \times m}$}
\KwResult{$\tilde{Z}_{i}\in\mathbb{R}^{m\times d}$}
$Z'_{i} \gets SCA(Z_i,X_i,S)$\;
$Z'_{i} \gets LN(Z'_{i}+Z_i)$\;
$\tilde{Z}_{i} \gets FFN(Z'_{i})$\;
$\tilde{Z}_{i} \gets LN(\tilde{Z}_{i}+Z'_{i})$\;
\end{algorithm}

\begin{algorithm}
\caption{SAT-Self at layer i}\label{alg:sat_self}
\KwData{$\tilde{Z}_{i}\in\mathbb{R}^{m\times d}$;$M\in\mathbb{1}^{m \times m}$}
\KwResult{$Z_{i+1}\in\mathbb{R}^{m\times d}$}
$Z''_{i} \gets SCA(\tilde{Z}_{i},\tilde{Z}_{i},M)$\;
$Z''_{i} \gets LN(Z''_{i}+\tilde{Z}_{i})$\;
$Z_{i+1} \gets FFN(Z''_{i})$\;
$Z_{i+1} \gets LN(Z_{i+1}+Z''_{i})$\;
\end{algorithm}

\begin{algorithm}
\caption{SAT-Block at layer i}\label{alg:sat_block}
\KwData{$X_{i}\in\mathbb{R}^{H_i\times W_i\times C_i}$;$Z_{i}\in\mathbb{R}^{m\times d}$;$S\in\mathbb{1}^{ H_i\times W_i \times m}$}
\KwResult{$Z_{i+1}\in\mathbb{R}^{m\times d}$; $X_{i+1}\in\mathbb{R}^{H_{i+1}\times W_{i+1}\times C_{i+}}$}
$X_i^{flat} \gets \text{Flatten}(X_i,dim=[0,1])$\;
$S_{i} = \text{Downsample}(S,shape=(H_i,W_i,m))$\;
$S_i^{flat} \gets \text{Flatten}(S,dim=[0,1])$\;
$\tilde{Z}_{i} \gets \text{SAT-Cross}(Z_i,X_i^{flat},S_i^{flat})$\;
$Z_{i+1} \gets \text{SAT-Self}(\tilde{Z}_{i},\tilde{Z}_{i},M)$\;
$X_{i+1} \gets g(X_i)$ \Comment*[r]{g a strided convolution}\
\end{algorithm}

\subsection{Details on SCAM}
Here are the detailed implementations for the SCAM Block (See Algorithm \ref{alg:scam_block}). For even finer details (Hyperparameters, practical details) see the provided Pytorch code.

\begin{algorithm}
\caption{SCAM-Block at layer j}\label{alg:scam_block}
\KwData{$X'_{j}\in\mathbb{R}^{H_j\times W_j\times C_i}$;$Y'_{j}\in\mathbb{R}^{H_j\times W_j\times 3}$; $Z'_{j}\in\mathbb{R}^{m\times d}$;$S\in\mathbb{1}^{ H_j\times W_j \times m}$}
\KwResult{$X'_{j+1}\in\mathbb{R}^{H_{j+1}\times W_{j+1}\times C_{j+1}}$;$Y'_{j+1}\in\mathbb{R}^{H_{j+1}\times W_{j+1}\times 3}$; $Z'_{j}\in\mathbb{R}^{m\times d}$}
$X'_{j}, Z'_{j} \gets SCAM(X'_{j},Z'_{j}, S) $\;
$X'_{j} \gets \text{Upsample}(X'_{j},upsize=2)$\;
$X'_{j+1}, Z'_{j+1} \gets SCAM(X'_{j},Z'_{j}, S) $\;
$X'_{j, RGB}, _  \gets SCAM(X'_{j+1},Z'_{j+1}, S) $\;
$Y'_j \gets \text{Upsample}(Y'_j, upsize=2)$\;
$Y'_{j+1} \gets Y'_j + X'_{j, RGB}$\
\end{algorithm}

\subsection{Losses}

Here, we complement Section 3.4 in the main paper and describe with more details the losses used for training \mname. We denote by G the SCAM-Generator, E the SAT-Encoder and D the PatchGAN discriminator.

\noindent For the\textbf{ GAN loss}, we use Hinge GAN loss~\cite{lim2017geometric} as follows: 
\begin{gather}
\begin{split}
        &\mathcal{L}_{\text{D,GAN}} = \mathbb{E}\left[ \text{max}(0,1-D(X,S))\right]\\ &+\mathbb{E}\left[ \text{max}(0,D(G(E(X,S),S),S)+1)\right] \\
\end{split}  \\
        \mathcal{L}_{\text{F,GAN}} = \mathbb{E}\left[D(G(E(X,S),S),S)\right] \quad . 
\end{gather}

\noindent For the \textbf{reconstruction loss}, we use the perceptual loss as in~\cite{Zhu_2020}. This uses a pretrained VGG network, and tries to make the intermediate feature maps between the input and the reconstructed input as close as possible. It is defined as:
\begin{equation}
     \mathcal{L}_{\text{VGG}} = \mathbb{E}\left[\sum_{i=1}^L\lVert F_i(X) - F_i(G(E(X,S))) \rVert_1 \right] \quad ,
\end{equation}
\noindent with $L$ the number of VGG hidden layers and $F_i$ the $i^{th}$ layer feature map of the VGG feature map.

We also use a $\mathcal{L}_1$ loss between the input and the reconstruction.

\begin{equation}
     \mathcal{L}_1 = \mathbb{E}\left[\lVert X -G(E(X,S)))  \rVert_1 \right] \quad ,
\end{equation}
\subsection{Metrics}
Here we give more details on the metrics we introduce: REIDSim and REIDAcc. Let's call $I^S$ the subject images, $I^B$ the background images and $I^{ST}$ the subject transfer images. Note that the subject images/background images couples are fixed to be able to compare on the same 
Now if REID is the REID network introduced in \cite{fu2021unsupervised}, we now have $X^S = REID(I^S)$, $X^B = REID(I^B)$ and $X^B = REID(I^B)$ the respective embeddings. Now we can compute the REID metric:
\begin{equation}
    REIDSim(I^S, I^{ST}) = \frac{1}{n}\sum_{k=1}^n \frac{(X^S_k)^TX^{ST}_k}{\lVert X^S_k\rVert\lVert X^{ST}_k\rVert}
\end{equation}

\begin{equation}
    REIDAcc(I^S, I^B, I^{ST}) = \frac{1}{n}\sum_{k=1}^n \mathbb{1}\left( \frac{(X^S_k)^TX^{ST}_k}{\lVert X^S_k\rVert\lVert X^{ST}_k\rVert}>\frac{(X^B_k)^TX^{ST}_k}{\lVert X^B_k\rVert\lVert X^{ST}_k\rVert} \right)
\end{equation}
With $n$ the number of subject transfer images we want to evaluate.

\subsection{Model characteristics} 

\begin{table}[t]
\begin{center}
		\begin{small}
\begin{tabular}{lrrr}
\toprule
Method & \# parameters $\downarrow$ & Training speed  $\downarrow$ & Inference speed $\downarrow$ \\
\midrule
SPADE & 109M & 95ms & 39ms\\
CLADE & 84M & \textbf{86ms} & \textbf{38ms} \\
SEAN-CLADE & 240M & 384ms & 83ms \\
SEAN & 265M & 280ms & 92ms \\
INADE & 85M & 179ms & 50ms \\

\mname & \textbf{95M} & 180ms & 61ms \\
\bottomrule
\end{tabular}
		\end{small}
	\end{center}
	\caption{\small{\textbf{Comparison} of model characteristics. Speeds are given in ms/samples. We evaluate on CelebAMask-HQ\cite{karras2018progressive} 20 semantic labels. \mname has $k=8$.}}
\label{tab:model_tech}
\end{table}

Thanks to the SCAM-Block, \mname results in 2x fewer number of parameters compared to SEAN (i.e., 95M vs 265M). SEAN is bigger than our model, as it comprises a big FFN for each style code at each block. Instead, with the \mname-Block, we have parameter sharing for every token in SCAM blocks, removing this constraint. Moreover, \mname trains faster and can infer values 50\% faster than SEAN. Measurements are made on a single NVIDIA RTX 3090.

\section{Additional experimental analysis}
\label{sup:sub:exps}

In this section, we display complementary experiments on the task of pose transfer. We also showcase some qualitative ablations of \mname.

\subsection{Pose Transfer}

Here, we illustrate qualitative results for the pose transfer task for \mname on the iDesigner~\cite{idesigner} dataset. 
Given two images, one being the style image $X_{Style}$ and another of $X_{Pose}$ being the pose reference. $S_{Pose}$ the associated segmentation mask. The goal of pose transfer is to generate an image matching the style of $X_{Style}$ with the semantics of $X_{Pose}$. We extract the latent codes $Z_{Style}$ with the SAT-Encoder. We then can generate $X_{PT} = G(Z_{Style},S_{Pose})$ the generated image that share the style of $X_{Style}$ and the pose of $X_{Pose}$.

\begin{figure*}[t!]
    \centering
    \includegraphics[width=\textwidth]{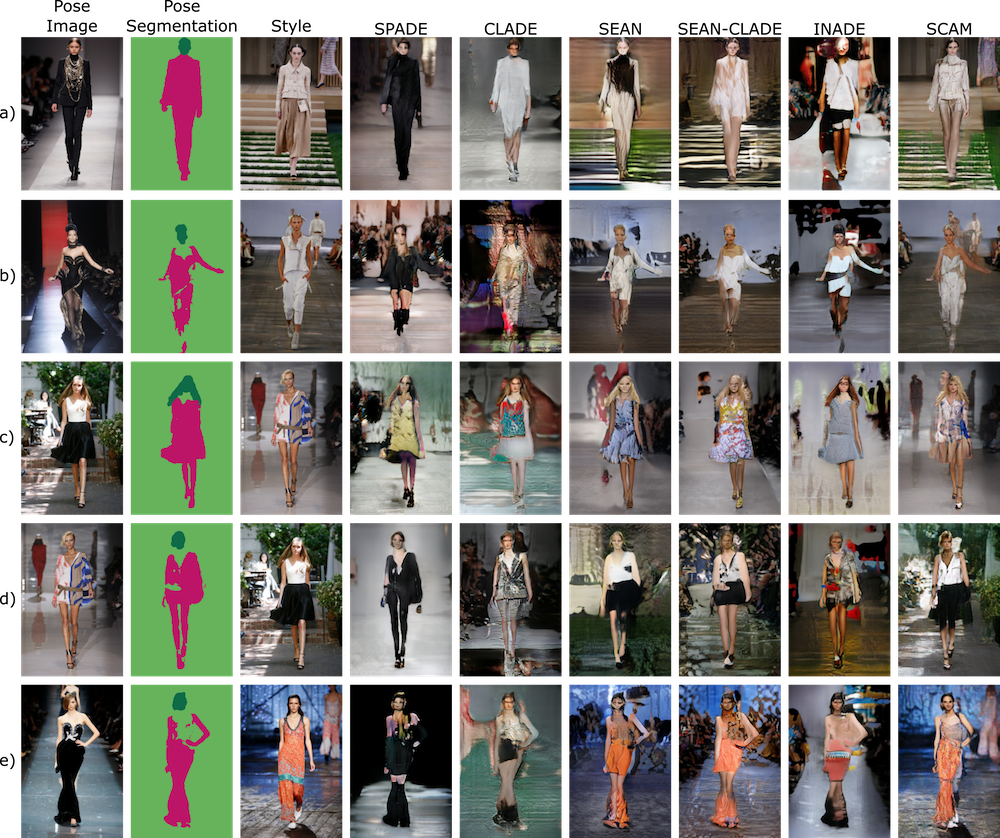}
     \caption{
     \small{\textbf{Pose Transfer in idesigner}. 
     We compare SCAM to competing methods on the pose transfer task.
     }}
    \label{fig:pose_transfer_idesigner}
\end{figure*}

In Figure \ref{fig:pose_transfer_idesigner}, we observe how SPADE, SEAN, SEAN++ and SCAM perform pose transfer. In this figure, we observe similar trends  that we observe with subject transfer.

\mname captures more details than the competing methods. This is particularly visible on the backgrounds of images a) b) and d), where the backgrounds on SCAM display more coherent outputs with the style reference image than other methods.

Similarly, we can observe that \mname presents better subject reconstruction like in a), where given the segmentation map, \mname generates a coherent subject. The skirt gets converted into pants because of the segmentation masks, but the texture remains the one of the style image. For SEAN and SEAN-CLADE, we do not see it generating coherent clothes that match the style of the style image.

\subsection{Ablations}
\begin{figure*}[t!]
    \centering
    \includegraphics[width=\textwidth]{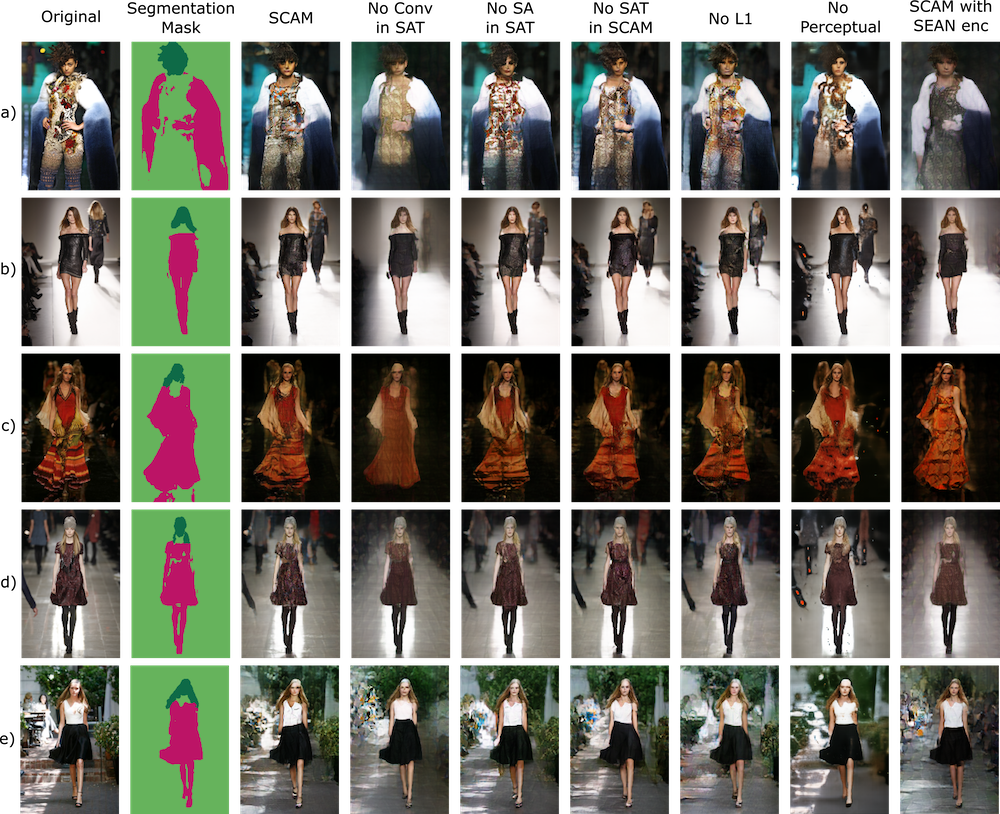}
     \caption{
     \small{\textbf{Qualitative ablations on idesigner}. 
     We perform the same ablations as in the quantitative ablations section of the main paper.
     }}
    \label{fig:qualitative_ablations}
\end{figure*}

In Figure \ref{fig:qualitative_ablations} we showcase some qualitative ablations to complement the main paper quantitative ablations. This additional visual ablation help us understand what each block of \mname brings perceptually.

We observe that when removing the convolutions (fourth column) in the SAT encoder, we end-up with a very simple texture that lacks details. This is particularly true in row c), where the dress texture is very repetitive and simplistic. This confirms our hypothesis that using information at multiple resolutions gives higher quality encodings.

If we remove self attention in SAT (fifth column), we observe that the reconstruction lack details in the coarser label, such as the background. For example, in row d), we can observe that the 2 person in the background features the same cloths colour where in SCAM we can see 2 different colours. Allowing latents from the same region interacting together in the encoder enables refining such details.

Removing the SAT in SCAM (sixth column) yields similar results. For instance, we observe in row b) that the background is less sharp than in \mname. The background person seems to be floating, whereas \mname has a more coherent interaction between the floor and the background person. Refining the latents in the SCAM-Generator also yields improves the semantic knowledge of SCAM.

When removing the $\mathcal{L}_1$ loss (seventh column) we have outputs similar to the ones in \mname (third column). However, there are some colourized artefacts, like in row a) (No L1). The $\mathcal{L}_1$ loss allows removing artefacts that arise from the perceptual loss.

When removing the perceptual loss (eighth column), we see some salt and pepper noise appearing in the output, like in row b). These artefacts are linked to the $\mathcal{L}_1$ loss and are stabilized by the perceptual loss.

When replacing the SAT encoder with the SCAM encoder (eighth column), we study the benefits of our encoder compared to the SEAN encoder. We observe that the SEAN Encoder fails to capture details in the background, and hence it limits the generator to only simplistic texture generation. This is particularly the case in row b) and d)  where the method totally misses the person in the background

\subsection{User Study}
\begin{figure*}[t!]
    \centering
    \includegraphics[width=\textwidth]{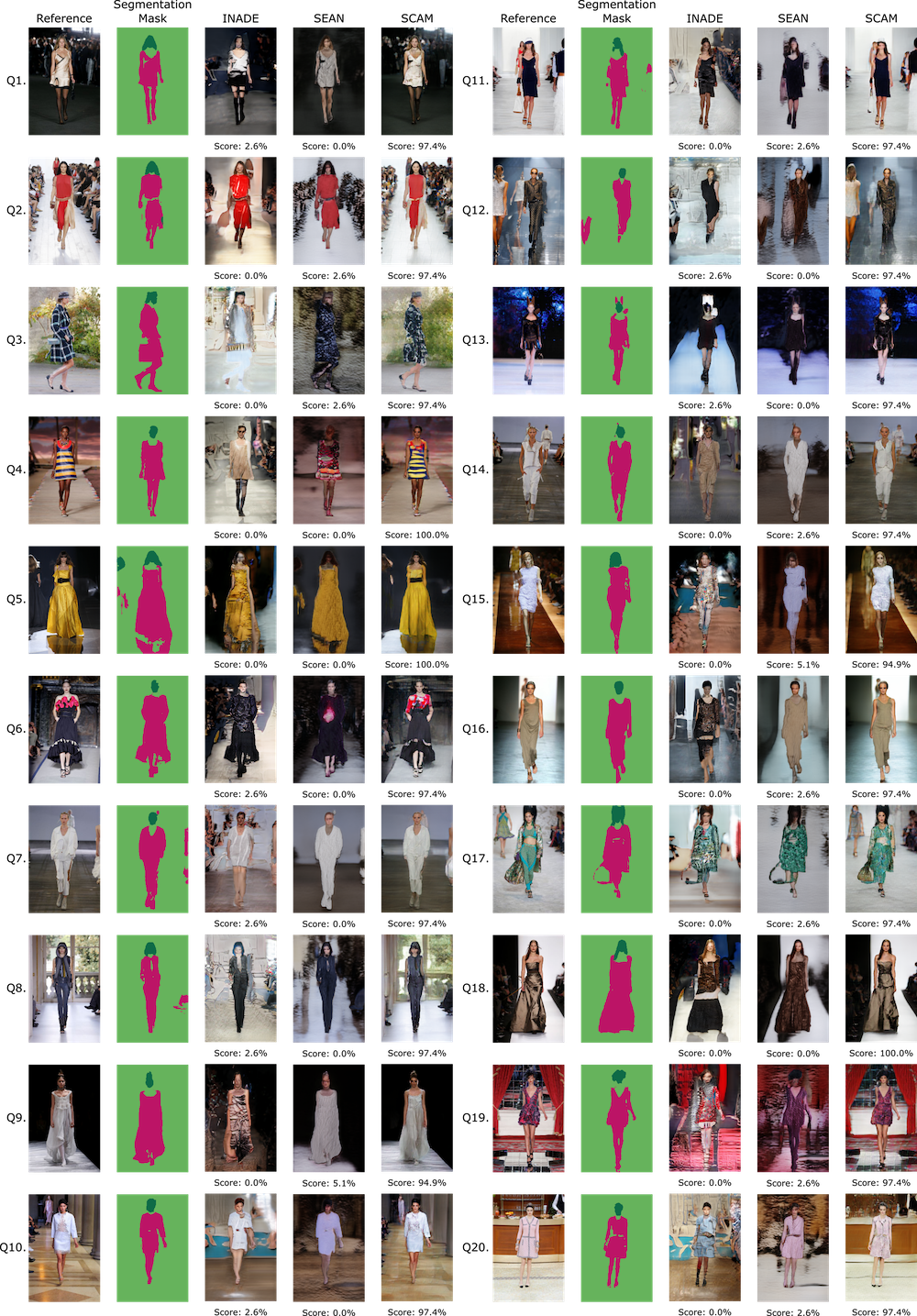}
     \caption{
     \small{\textbf{User study on reconstruction}. 
     We provide here the samples that were used to evaluate the reconstruction quality on SCAM, SEAN and INADE in a user study. Note that we provide here the segmentation mask for reference but it was not provided to users.
     }}
    \label{fig:user_study_reco}
\end{figure*}

\begin{figure*}[t!]
    \centering
    \includegraphics[width=\textwidth]{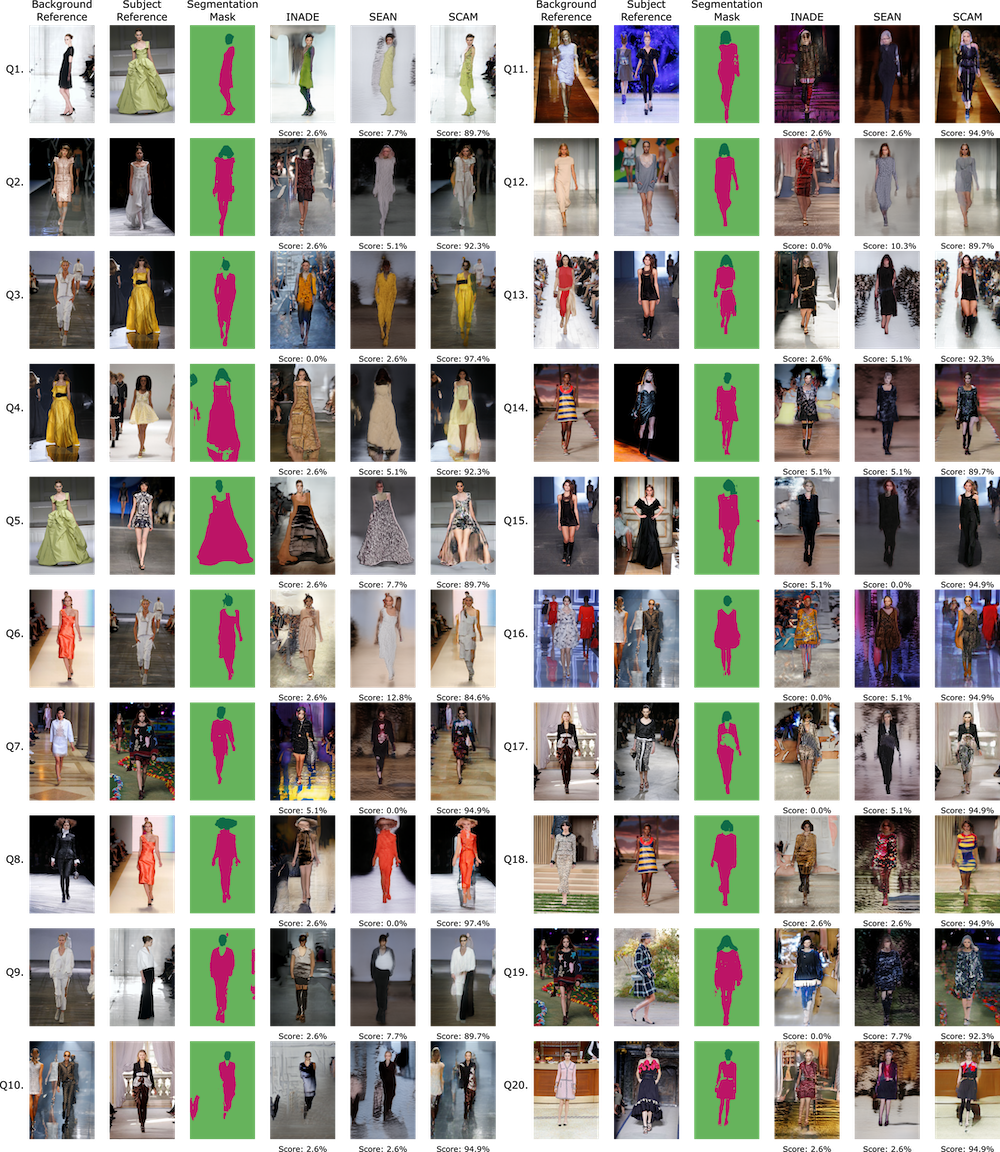}
     \caption{
     \small{\textbf{User study on subject transfer}. 
     We provide here the samples that were used to evaluate the subject quality on SCAM, SEAN and INADE in a user study. Note that we provide here the segmentation mask for reference but it was not provided to users.
     }}
    \label{fig:user_study_st}
\end{figure*}

We provide in Figure \ref{fig:user_study_reco} and \ref{fig:user_study_st} the samples we used for the user study and the score for each individual question.

\subsection{Additional visual results}
\begin{figure}[h!]
    \centering
    \includegraphics[width=\textwidth]{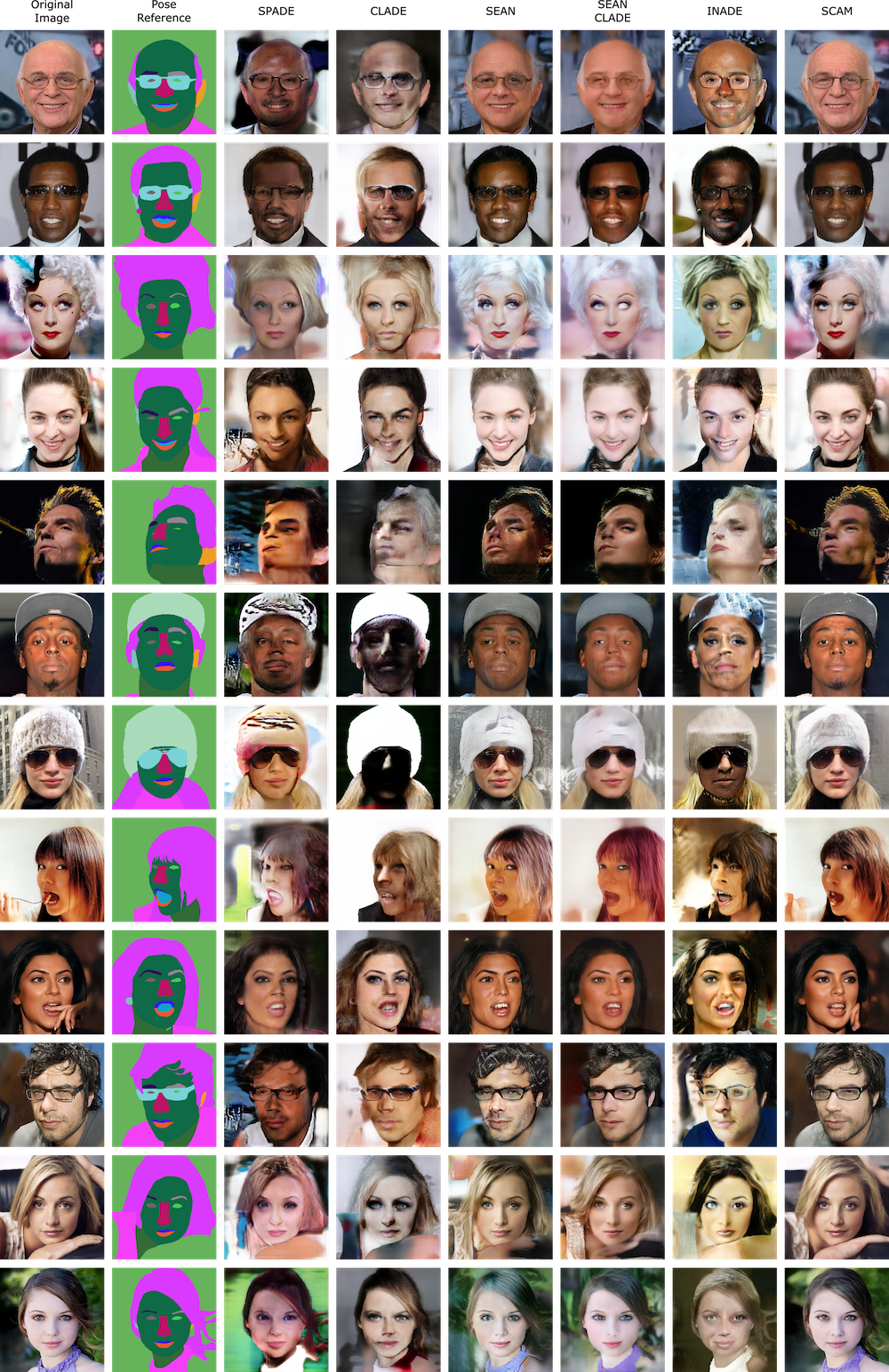}
     \caption{
     \small{\textbf{Reconstruction on CelebAMask-HQ}. 
    }}
    \label{fig:reco_celeb_supp}
\end{figure}

\begin{figure}[h!]
    \centering
    \includegraphics[width=\textwidth]{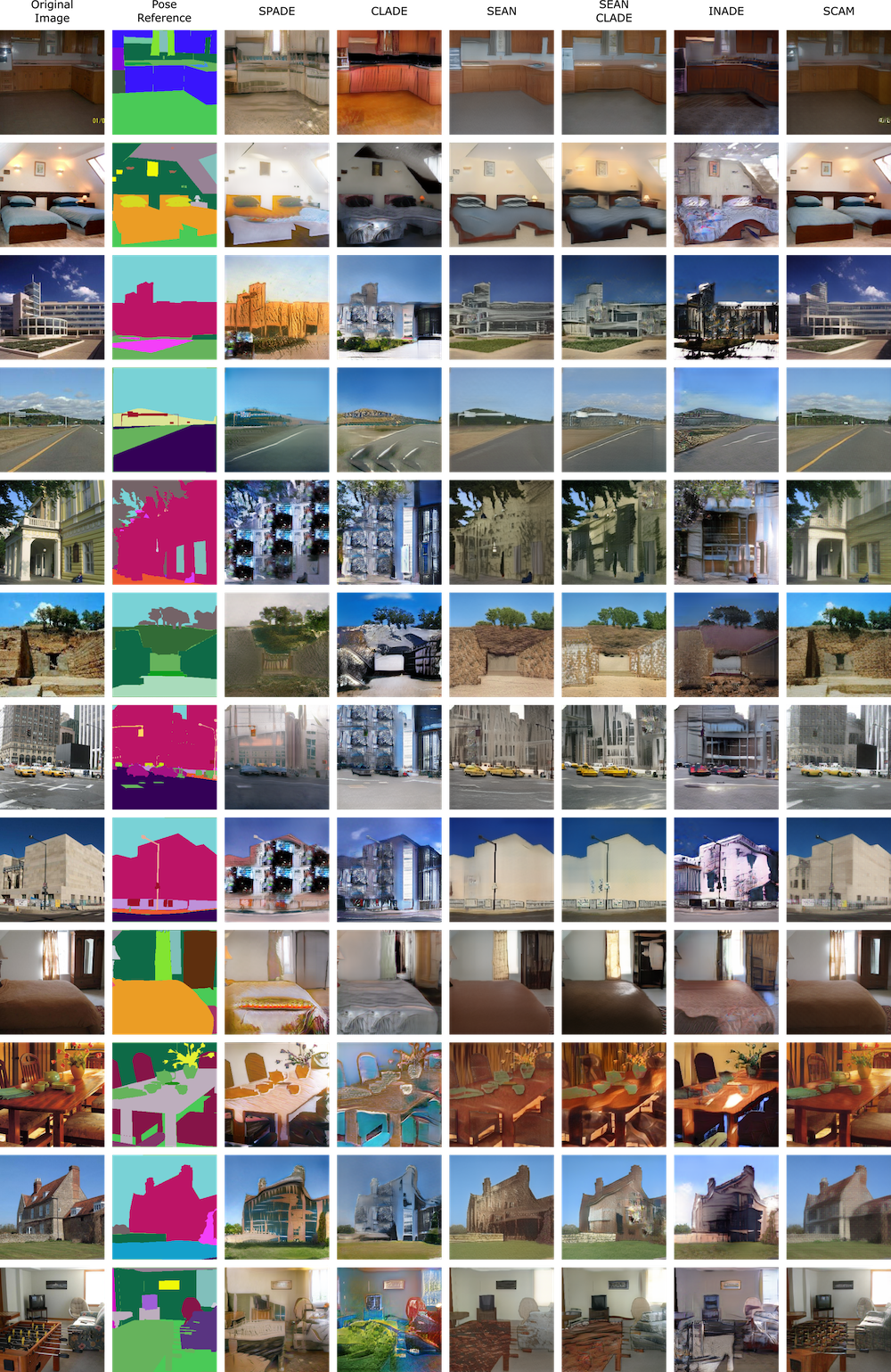}
     \caption{
     \small{\textbf{Reconstruction on ADE20K}. 
     }}
    \label{fig:reco_ade20k_supp}
\end{figure}

We provide in Figure \ref{fig:reco_celeb_supp} and \ref{fig:reco_ade20k_supp} additional results on CelebAMask-HQ and ADE20K.

\end{document}